\UseRawInputEncoding
\documentclass[conference]{IEEEtran}

\usepackage{cite}
\usepackage{amsmath,amssymb,amsfonts}
\usepackage{algorithmic}
\usepackage{graphicx}
\usepackage{textcomp}
\usepackage{xcolor}
\usepackage[ruled,linesnumberedhidden]{algorithm2e} 

\usepackage{caption}
\usepackage{subcaption}

\usepackage[textsize=tiny]{todonotes}
\usepackage{comment}

\def\BibTeX{{\rm B\kern-.05em{\sc i\kern-.025em b}\kern-.08em
    T\kern-.1667em\lower.7ex\hbox{E}\kern-.125emX}}

\usepackage[hyphens]{url}
\usepackage{hyperref}
\usepackage[hyphenbreaks]{breakurl}

\begin{document}
\title{End-to-End Verifiable Decentralized \\Federated Learning}

\author{
    \IEEEauthorblockN{
    Chaehyeon Lee\IEEEauthorrefmark{1},
    Jonathan Heiss\IEEEauthorrefmark{2},
    Stefan Tai\IEEEauthorrefmark{2},
    and James Won-Ki Hong\IEEEauthorrefmark{1}
}

\IEEEauthorblockA{
    \IEEEauthorrefmark{1}Department of Computer Science and Engineering, POSTECH, Pohang, Korea
    \\\IEEEauthorrefmark{2}Information Systems Engineering, TU Berlin, Berlin, Germany
}
\IEEEauthorblockA{
    Email: chlee0211@postech.ac.kr, j.heiss@tu-berlin.de, tai@tu-berlin.de, jwkhong@postech.ac.kr\\
}
}

\maketitle

\begin{abstract}
Verifiable decentralized federated learning (FL) systems combining blockchains and zero-knowledge proofs (ZKP) make the computational integrity of local learning and global aggregation verifiable across workers.
However, they are not end-to-end: data can still be corrupted prior to the learning. 
In this paper, we propose a verifiable decentralized FL system for end-to-end integrity and authenticity of data and computation extending verifiability to the data source. 
Addressing an inherent conflict of confidentiality and transparency, we introduce a two-step proving and verification (2PV) method that we apply to central system procedures: a registration workflow that enables non-disclosing verification of device certificates and a learning workflow that extends existing blockchain and ZKP-based FL systems through non-disclosing data authenticity proofs. 
Our evaluation on a prototypical implementation demonstrates the technical feasibility with only marginal overheads to state-of-the-art solutions.

\end{abstract}
\begin{IEEEkeywords}
federated learning, blockchain, off-chain computations, verifiability, zero-knowledge proofs, zokrates
\end{IEEEkeywords}

\section{Introduction}
\label{sec:introduction}
Federated learning (FL)~\cite{mcmahan_communication_2017} helps to overcome confidentiality and network challenges of machine learning in distributed settings like the Internet of Things (IoT) by separating the responsibilities of learning and aggregation~\cite{nguyen_FLIoTSurvey_2021}. 
The learning is executed on distributed workers to keep confidential learning data protected. 
In each learning cycle, only updated learning parameters are shared and aggregated into a global model that represents the collective learning progress and is returned to the workers for the next learning cycle. 
However, the splitting and outsourcing of responsibilities in the FL process introduces security issues in distributed environments where parties do not trust each other or a central aggregator, and hence, a strong need for the verification of FL tasks.

Decentralized FL through blockchains removes the need for a central aggregator that may fail or corrupt the result aggregation unnoticeably. 
Global tasks are executed through smart contracts in a transparent and tamper-resistant manner allowing participating nodes to verify aggregation correctness by participating in the consensus protocol.
As such, blockchain-based FL has been applied in different domains and associated challenges addressed in a large body of literature~\cite{zhao_AuthFLVehicles_2022,wang_authFLVehicleEnergy_2021,safa_energyTradeFL_2023,Ouns_federatedGrids_2022,Hierarical_fedLearn_healthcare_2021,FedLearn_Healthcare_2018}. 
While blockchains add a first notion of verification to decentralized FL systems, they are unsuitable for executing the local learning procedures due to confidentiality and scalability constraints.  

To prevent model poisoning~\cite{xia_poisoningSurvey_2023}, verifiable decentralized FL systems have been proposed that advance blockchain-based FL through verifiable off-chain computation~(VOC) where local model updates are executed using zero-knowledge proofs (ZKP)~\cite{advancing_heiss_2022,sedlmeir_2022_fairness,ZKP-PracFed_xing_2023,bv-ICVs_smahi_2023}. 
By verifying these ZKPs on the blockchain, the computational integrity of local learning becomes verifiable throughout the network without disclosing confidential learning inputs. 
However, VOC-based FL systems are not end-to-end: Learning data can still be corrupted by workers prior to the learning.

End-to-end verifiable decentralized FL systems add a third notion of verification of data sources and data authenticity, thereby protecting against Sybil attacks~\cite{douceur2002sybil} and data poisoning~\cite{xia_poisoningSurvey_2023}. 
This is especially relevant in IoT settings where the learning cannot be executed on constraint IoT devices but is outsourced to an intermediary node run by the same worker. 
For example, in smart energy applications~\cite{safa_energyTradeFL_2023,Ouns_federatedGrids_2022}, learning is outsourced from smart meters to workstations or in smart healthcare applications~\cite{FedLearn_Healthcare_2018,Hierarical_fedLearn_healthcare_2021}, from medical devices to hospital facilities. 

The verification of authenticity proofs of learning data and certificates of device identities on the blockchain, however, encounters an inherent conflict of transparency and confidentiality.
Signatures created on the learning data do not apply to the updated learning parameters after the learning and their verification on the blockchain discloses confidential learning data to the blockchain network.
Furthermore, the verification of device certificates requires the disclosure of certificate attributes like the device public key which represent security leaks and must be protected, e.g., to prevent key search attacks. 

Addressing this problem, in this paper, we suggest a first \textit{end-to-end verifiable decentralized FL system} that makes the integrity and authenticity of data and computation verifiable, from certified edge devices to the blockchain’s secure parameter storage. 
Our system builds upon state-of-the-art approaches leveraging blockchains for aggregation tasks and zkSNARKs for local learning tasks, e.g.,~\cite{advancing_heiss_2022}, and extends them through integrated authenticity proofs of certified sensor devices. 
For that, we make three individual contributions:

\begin{enumerate}
    \item First, we propose a system model that extends VOC-based FL systems through an integrated perspective of certified sensor devices acting as data sources. We derive verification objectives from a motivational healthcare use case and refine the problem of conflicting properties of confidentiality and transparency. 
    \item Second, we present a two-step proving verification (2PV) procedure realized in a registration and a learning workflow as the two central procedures for our system. The registration workflow onboards certified sensor devices by verifying their device certificates without disclosing confidential attributes. The learning workflow provides local mode updates by verifying both, the integrity of the learning and the authenticity of the learning data without disclosing them. 
    \item Third, we evaluate our system proposal through a prototypical implementation using ZoKrates~\cite{eberhardt_tai_2018} and initial experimentation of both workflows. The results demonstrate that our solution only adds a marginal overhead to previous comparable implementations~\cite{advancing_heiss_2022} but achives end-to-end verifiability. 
\end{enumerate}

In the remainder of this paper, we first provide background information in Section~\ref{sec:background}, then discuss related work in Section~\ref{sec:relatedWork}, and present our system model in Section~\ref{sec:application_model}.
Based on that, we describe both workflows in more detail in Section~\ref{sec:systemDesign}, present our implementation in Section~\ref{sec:Implementation}, and the evaluation in Section~\ref{sec:Evaluation}. 
We discuss remaining issues in Section~\ref{sec:Discussion} and finally conclude in Section~\ref{sec:Conclusion}.

\section{Background}
\label{sec:background}
We start describing the foundations of federated learning (FL) and zkSNARKs as central concepts of our system design.

\subsection{Federated Learning}
\label{sec:fedlearn}
FL offers a confidentiality-preserving and network-efficient architectural framework for machine learning tasks in distributed settings.
As depicted in Figure~\ref{fig:fed_learn}, a single global model $GM$ is iteratively trained. This training occurs through a large set of distributed \textit{workers}, each locally executing learning tasks on confidential learning data $LD$ in each cycle.
The resulting local model updates $LM$ consisting of learning parameters, such as weights $w$ and biases $b$, are then integrated into a global model $GM$ through an \textit{aggregator} that, in decentralized FL systems, can be executed on blockchains. 
The global model represents the collective learning insights of the network and is returned to the workers for the next learning iteration. 
This mechanism enables each worker to benefit from the collective training insights garnered by other workers while safeguarding potentially sensitive inputs from disclosure and reducing network loads~\cite{fedLearn_originalPaper_2016}.

\subsection{ZkSNARKs and ZoKrates}
\label{sec:zksnarks}
Zero-knowledge succinct non-interactive arguments of knowledge (ZkSNARKs) represent a specific category of non-interactive zero-knowledge proofs (NIZK) known for their compact proof sizes and efficient verification times. 
The zkSNARKs procedure can be conceptualized through three main operations: 

\begin{figure}[h]
    \centering
    \includegraphics[width=1\columnwidth]{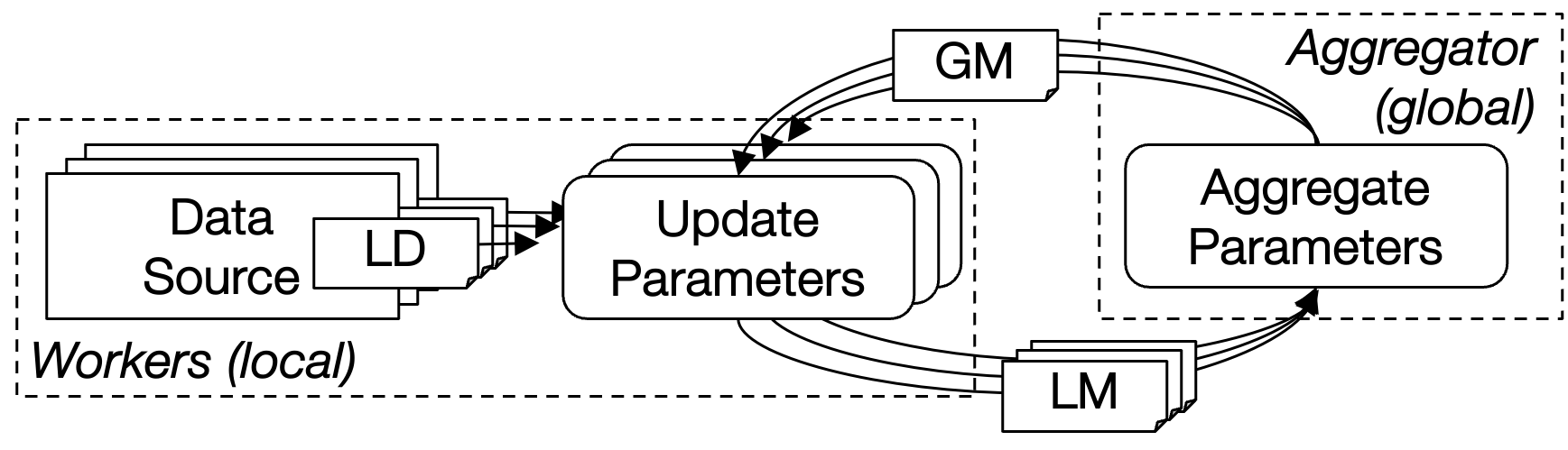}
    \caption{Federated Learning Overview}
    \label{fig:fed_learn}
\end{figure}

\begin{itemize}
\item The \textit{setup} ($setup(ecs, srs) \rightarrow (ZK_{prov},ZK_{verif})$) generates a public asymmetric key pair derived from the executable constraint systems ($ecs$). 
The $ecs$ encodes the program logic in a provable representation, and a circuit-dependent structured reference string ($srs$) is used. 
Both proving and verification keys ($ZK_{prov},ZK_{verif}$) are tied to the $ecs$. 
This process assumes a secure disposal of the $srs$ to thwart any attempts at producing fake proofs.

\item The \textit{proving} ($P(ecs, x, x', w, ZK_{prov}) \rightarrow \pi$) occurs in two steps: 
Firstly, a witness $w$ is created by executing the $ecs$ on the public inputs $x$ and the private inputs $x'$. 
The tuple ($x$, $x'$) constitutes the proof arguments. 
The witness $w$ signifies a valid variable assignment for the $ecs$ inputs. 
Subsequently, the proof $\pi$ is generated from the witness using the proving key.

\item The \textit{verification} ($V(\pi, x, ZK_{verif}) \rightarrow \{0, 1\}$) takes place on-chain, evaluating the proof $\pi$ and the public inputs $x$ using the verification key $ZK_{verif}$.
\end{itemize}

ZkSNARKs are particularly well-suited for applications in Verifiable Off-Chain Computing (VOC) where the verification is executed on the blockchain. 
VOC helps to overcome the privacy and scalability limitations of blockchains by offloading computation without compromising the blockchain's integrity. 
\textit{ZoKrates}~\cite{eberhardt_tai_2018} is a language and toolbox that facilitates the development of zkSNARKs-based VOC for Ethereum~\cite{wood2014ethereum}. 
The ZoKrates language provides a convenient interface for developers to define proving logic. The ZoKrates toolbox supports the compilation into $ecs$, the setup, the witness computation, and proof generation, as well as the creation of verifier contracts in Solidity. 

\section{Related Work}
\label{sec:relatedWork}
We position our contributions within related research for verifiable and decentralized federated learning (FL)~\cite{zhang_towardsVFL_2022} comprising blockchains for verifiable global tasks, zero-knowledge proofs (ZKP) for verifiable learning tasks, and authentication schemes against data poisoning.

\textbf{Verifiable Aggregation with Blockchain:}
Blockchain-based FL has been proposed in different domains like smart energy grid~\cite{safa_energyTradeFL_2023,Ouns_federatedGrids_2022}, smart healthcare~\cite{FedLearn_Healthcare_2018,Hierarical_fedLearn_healthcare_2021}, or smart homes~\cite{FedLearn_SmartHome_2019} and been researched under different objectives. 
Protection against \textit{leakage of confidential information} from learning parameters is important for blockchain-based FL where local models become accessible by the blockchain network.  
Approaches leveraging differential privacy techniques applied to the learning data have been proposed in~\cite{sedlmeir_2022_fairness,lu_blockchainFLIoT_2019,FedLearn_Blockchain_Privacy_2021}. Besides, \cite{zhou2024vdfchain} addresses Byzantine attacks by employing a trusted mechanism based on polynomial commitment, ensuring correctness of the aggregation model.
As another challenge, \textit{fairness} refers to an equal and balanced contribution of the workers and has been addressed in~\cite{trustworthyFedLearn_2021,lyu_fairnessFLBlockchain_2019,sedlmeir_2022_fairness} through accountability and incentive mechanisms. 
Furthermore, high transaction costs that incur through blockchain-caused overheads present a problem in constrained environments. 
To this end, approaches have been proposed advocating more efficient FL implementations~\cite{li_chen_liu_huang_zheng_yan_2020}, using permissioned blockchains implementing lightweight consensus protocols by design~\cite{FabricFL}, or executing the aggregation off-chain using ZKPs to preserve computational integrity~\cite{zkfl_knottenbelt_2023}.
While these approaches represent important lines of work, we set a different focus in this paper and consider such privacy, fairness, and performance features complementary to our system.

\textbf{Verifiable Local Learning with ZKPs:}
Blockchain-based FL has been advanced by executing local learning as verifiable off-chain computations (VOC) with zkSNARKs, e.g., in~\cite{advancing_heiss_2022,sedlmeir_2022_fairness,ZKP-PracFed_xing_2023,bv-ICVs_smahi_2023}
In~\cite{advancing_heiss_2022}, blockchain-based FL has been extended through zkSNARKs-based local learning. 
A similar system has been proposed in~\cite{sedlmeir_2022_fairness} that additionally leverages blockchains for worker incentives and applies a differential privacy scheme to prevent information leakage.
In~\cite{ZKP-PracFed_xing_2023}, another VOC-based FL system is proposed where inference from local models is prevented through the secure sum protocol.
The authors of~\cite{bv-ICVs_smahi_2023} propose a similar system for mobility environments where connected vehicles represent learning nodes and zkSNARKs create local models verifiable on a domain-specific blockchain system for aggregation.
In this work, we conceptually extend these approaches through the verification of authenticity proofs and device identities using the prototype of~\cite{advancing_heiss_2022} as our reference implementation.

\textbf{Verifiable Data Sources and Data Authentication}
Authenticity proofs in blockchain-based FL systems have been proposed to establish accountability and mitigate poisoning and Sybil attacks. 
In~\cite{fan_AuthFedLean_2023}, an anonymous authenticity scheme based on a Public Key Infrastructure (PKI) is proposed for workers in a blockchain-based FL system that leverages allows for holding workers accountable without revealing their identities on the blockchain. 
Similar approaches are proposed in~\cite{wang_authFLVehicleEnergy_2021} and~\cite{zhao_AuthFLVehicles_2022} for vehicle-based workers.
These approaches create authenticity proofs on computed learning parameters allowing for model and data poisoning despite the achieved accountability. Instead, in this paper, we prove the authenticity of learning data, however, without disclosure.

\textbf{End-to-End Integrity Workflows in dApps:}
Outside the FL domain, end-to-end workflows have been proposed that leverage authentication schemes and zkSNARKs for data provisioning in blockchain-based decentralized applications (dApp)~\cite{heiss_trustworthypreprocessing_ICSOC2021,park_ziraffe_2020,Zhigou_zk-AuthFeed_2023}.
In~\cite{heiss_trustworthypreprocessing_ICSOC2021} trustworthy pre-processing in sensor data provisioning workflows has been proposed for blockchain-based IoT applications using zkSNARKs and TEEs. 
Wan et al.~\cite{Zhigou_zk-AuthFeed_2023} propose a zkSNARKs extension for signature-based data authentication applied to compute off-chain data in a manner that is verifiable on smart contracts.
Similarly, Park et al. propose Ziraffe~\cite{park_ziraffe_2020} as an approach for authenticating web-based data sources using zkSNARKs.
Such end-to-end approaches have been applied to verify credentials of dApp users~\cite{heiss_verifCredentials_2022} where data sources are trustworthy identity issuers or to verify carbon footprints~\cite{heiss_verifCarbAcc_2023} where data sources are certified sensor devices.
From a dApp perspective, we introduce data provisioning workflows to verify device certificates and local learning in FL dApps.

\section{Model}
\label{sec:systemModel}
In this section, we first derive verifiability objectives from a motivation scenario, then we provide an overview of the system architecture and finally refine our problem statement.

\subsection{Motivating Scenario and Verifiability Objectives}
\label{sec:secenario}
In smart healthcare applications, e.g., for remote patient tracking~\cite{FedLearn_Healthcare_2018,Hierarical_fedLearn_healthcare_2021}, machine learning can help predict a patient’s health state. 
The learning process takes health measurements, e.g., blood pressure, heartbeats, or temperature, as inputs and returns prediction classes like good, fair, critical. 
Such data is collected from medical devices distributed across hospitals, doctor’s offices, and patient’s homes, e.g., in cases of immobile patients. 
Medical devices undergo a strict certification procedure before they can be bought on the market.

As a problem for such smart healthcare applications, these measurements contain highly confidential information about a patient's health that fall under regulatory procetions like the Health Insurance Portability and Accountability Act (HIPAA)~\cite{HIPPA_2004}. 
Through federated learning, this health data can be kept under the control of the learning nodes, i.e., the patient or the responsible hospital, and must not be shared externally.

Due to the criticality of the prediction, it is of utmost importance that the learning procedure is executed correctly. 
Verifiable FL~\cite{zhang_towardsVFL_2022} helps to prevent corruption from faulty processes or malicious behavior. 
Verifiability objectives must be taken from the perspective of the affected parties.  
Consequently, we propose the following design objective for an end-to-end verifiable decentralized FL system. 
\begin{enumerate}
    \item [Obj1] The workers should be able to verify that the aggregation of the local model updates into a global model has not been corrupted to exclude aggregation attacks. 
    \item [Obj2] The workers should be able to verify the computational integrity of the local model updates to exclude model poisoning. 
    \item [Obj3] The workers should be able to verify that the data applied for the local learning updates is of authenticity and originates only from the expected certified devices to exclude data poisoning through workers. 
\end{enumerate}

\begin{figure}[h]
    \centering
    \includegraphics[width=1\columnwidth]{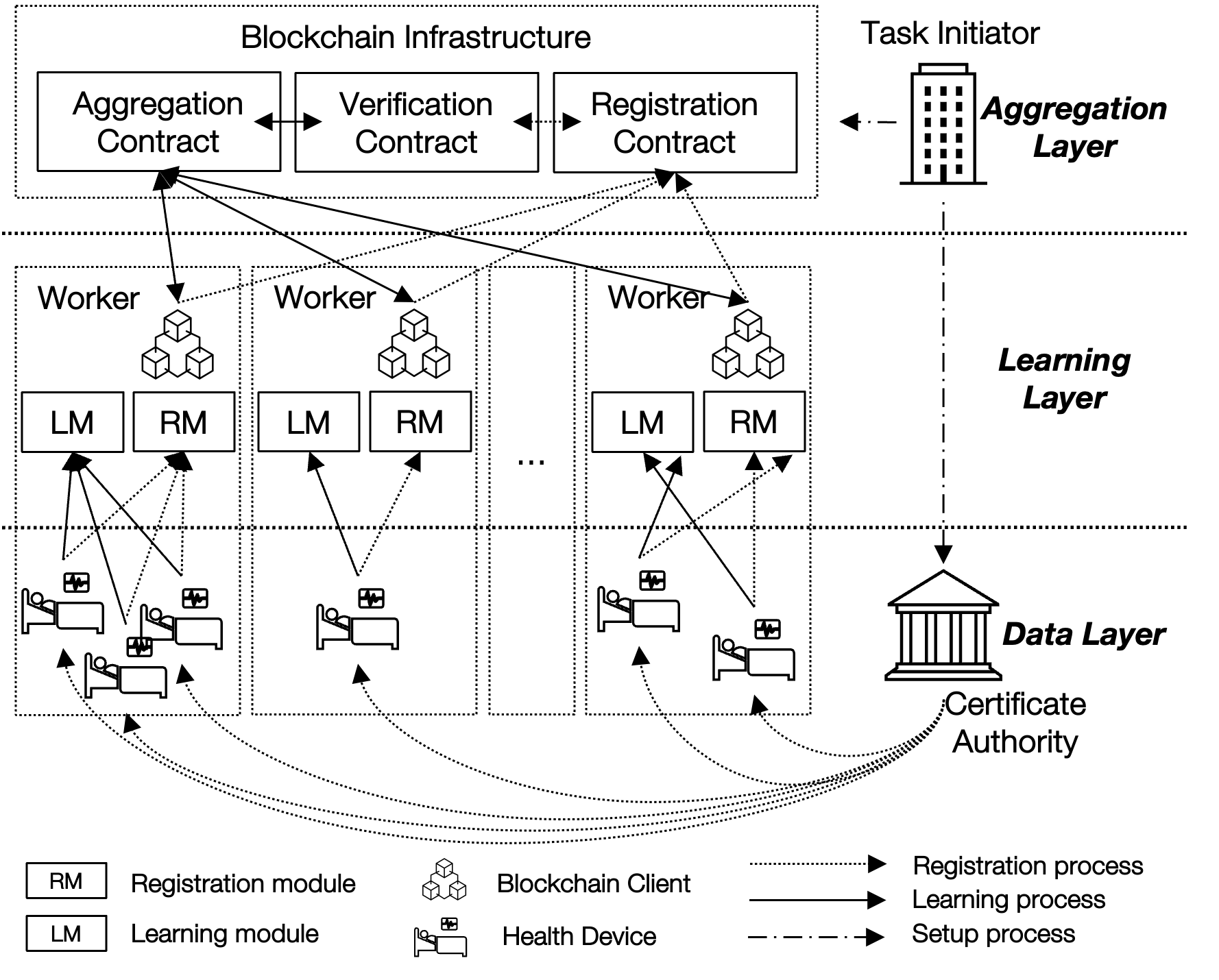}
    \caption{Overall System Architecture}
    \label{fig:architecture}
\end{figure}
\setlength{\textfloatsep}{5pt}

\subsection{System Overview}
\label{sec:application_model}
Building upon the state-of-the-art in verifiable decentralized FL and aiming toward the previous objectives, we now introduce the general system model that consists of three layers.

\subsubsection{Aggregation Layer}
The aggregation layer contains system components for executing global FL tasks that are set up through the task initiator.
To achieve \textit{obj1}, these components are executed on a blockchain infrastructure as smart contracts. 
The \textit{registration contract} manages the device identities. 
As a registration condition for a device, the smart contract verifies that they are certified under the same public key infrastructure (PKI).
The \textit{aggregation contract} contains the logic for aggregating the local learning parameters into a global model. 
As a condition for these local models to be aggregated, the integrity of the local learning procedure and the authenticity of the data need to be verified. 
The \textit{verification contract} provides verification services to the aggregation and the registration contract. 
It is responsible for verifying the proofs provided through the learning and registration workflows.

\subsubsection{Learning Layer}
The learning layer contains components for the local model updates that are executed on the workers' local infrastructures. Workers are distrusted for local and global operations and they may execute data and model poisoning and corrupt aggregation. 
To protect against model poisoning and fulfill \textit{Obj2}, we assume that relevant local tasks are executed through zkSNARKs as described in Section~\ref{sec:zksnarks}. 
Applying them as verifiable off-chain computations (VOC), the local model updates become verifiable on the blockchain without disclosing confidential inputs.
Each worker hosts two zkSNARK-enabled system components: 
The \textit{registration module} contains all logic to create proofs that allow for onboarding valid devices without disclosing certificate attributes.
The \textit{learning module} contains all the logic to create proofs of data authenticity and integrity of the learning procedure. 
Furthermore, workers connect to the smart contracts through a locally hosted blockchain client. 

\subsubsection{Data Layer}
As a distinguishing feature of our system, we separate the data layer from the learning layer. 
The data layer contains data sources that are accessible to the workers and generate learning data for the local model updates. 
This data contains confidential information that must not be disclosed to parties other than the associated workers. 
We assume these sources to be \textit{certified sensor devices} like smart meters, medical devices, or onboard devices that undergo a rigorous certification procedure including, for example, gauging, sealing, and securely installing the device. 
Certificates are managed in a public key infrastructure (PKI) and are issued through a \textit{certificate authority} (CA) that attests to the well-functioning and security of the devices.

\subsection{Problem Statement}
\label{sec:problem_statement}
Through zkSNARKs-based learning and blockchain-based aggregation, the integrity of computations in the FL process can be verified through the workers and respective attacks mitigated. 
However, achieving \textit{obj3} is hindered by an inherent conflict of confidentiality and verifiability.
We identify two specific challenges:

\begin{enumerate}
    \item [Ch1] To allow workers to verify that the expected data has been used for the local model updates, the data authenticity must be verified on the blockchain. 
    Authenticity proofs, e.g., through cryptographic signatures, require the data to be revealed during verification. 
    If done on the blockchain, confidential information contained in the learning data is revealed to the blockchain network. 
    
    \item [Ch2] The verification of device certificates helps to build trust in the learning data. 
    However, certificate attributes like the device public key must not become globally accessible to third parties as they may allow for insights into the internals of the device and with that may be used by external attackers. 
\end{enumerate}

\section{System Design}
\label{sec:systemDesign}

\begin{figure*}[t]
    \centering
    \includegraphics[width=0.85\textwidth]{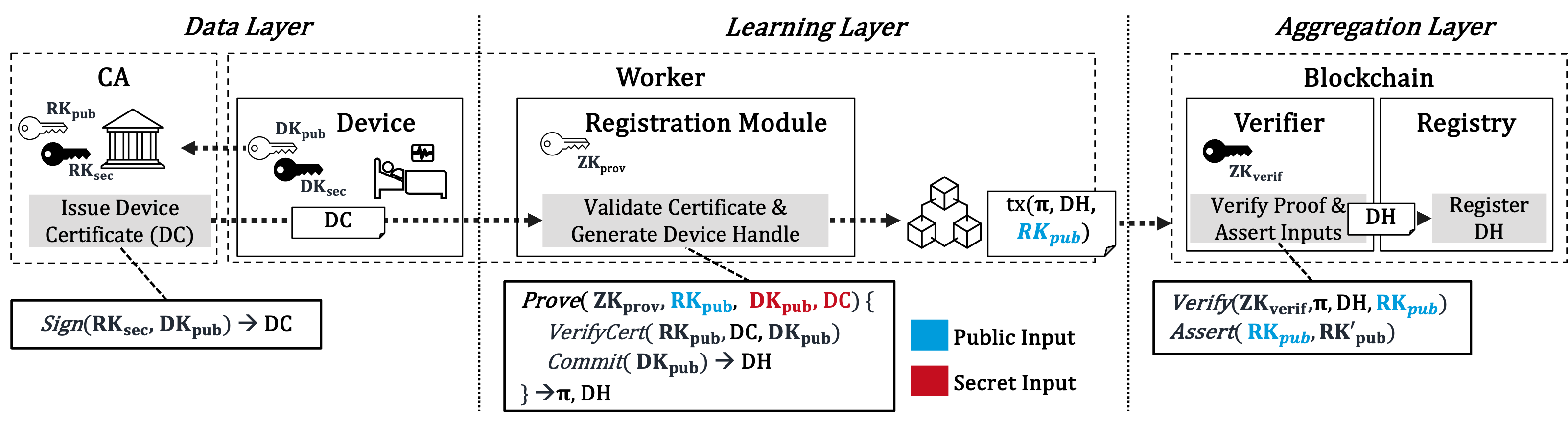}
    \caption{One-time \textbf{Registration Workflow} for Non-disclosing Device Certificate Verification.}
    \label{fig:registration_workflow}
\end{figure*}

Addressing the previously identified challenges, we present a \textit{two-step proving and verification} (2PF) procedure that is applied to realize two workflows that integrate into the previous system model:
a registration workflow for device identities addressing \textit{ch2} and a learning workflow for locally updated learning parameters addressing \textit{ch1}.
The 2PF procedure involves three steps. 

\begin{enumerate}
    \item The \textit{attestation} is executed on the data layer. A trusted party, i.e., the certificate authority or the certified sensor device, attests to confidential input data creating an authenticity proof verifiable through the worker nodes. 
    
    \item The \textit{proving} is executed on the learning layer. The distrusted workers create proofs of registration or learning depending on the workflow, each verifiable on the blockchain. These proofs contain the verification of the authenticity proofs from the data layer, thereby preventing disclosure of confidential inputs on-chain.
    
    \item The \textit{verification} is executed on the aggregation layer. The verifier contract verifies the submitted proofs from the workers and, if successful, provides the registration or learning outcome to the respective contract. 
\end{enumerate}

\subsection{System Initialization}
To set up the system, the task initiator creates the necessary artifacts and deploys the smart contracts. 
The artifacts provisioning to the workers is described in Section~\ref{sec:Discussion}.

To initiate the \textit{registration contract}, the task initiator obtains the PKI’s root key ($RK'_{pub}$) from the CA and anchors it into the contract.
The root key is required to validate that only the device certificates under the expected PKI are accepted and only corresponding device handles ($DH$) are registered.

For the \textit{aggregation contract}, the task initiator implements the aggregation strategy, e.g., averaged-based~\cite{nilsson_fedAveraging_2018}, and creates the initial learning parameters that will be input to the first learning cycle and then updated in each iteration. 

For the \textit{verification contract}, the task initiator creates one proving and verification key pair ($ZK_{prove}, ZK_{verif}$) for the registration workflow and one for the learning workflow by executing the zkSNARK setup respectively.
It specifies the proving logic executed in the registration and learning module respectively, compiles it into a $ecs$, and then executes the setup as defined in Section~\ref{sec:zksnarks}. 
The verification key $ZK_{verif}$ is submitted to the verification contract and, with that, the registration logic bound to the proving key $ZK_{prove}$ and $ecs$.

\subsection{One-Time Registration Workflow}
\label{sec:registration_workflow}
Successful device registration represents the prerequisite for participation in the federated learning system.
Consequently, the registration workflow is executed \textit{once} for each sensor device that contributes input data to the learning procedure.
As depicted in Figure~\ref{fig:registration_workflow}, it starts at the certificate authority (CA) and ends at the registration contract. 

\subsubsection{Attestation}
The CA issues the device certificate ($DC$) by signing the device's public key ($DK_{pub}$) using the root secret key ($RK_{sec}$) such that $Sign(RK_{sec}, DK_{pub}) \rightarrow DC$.
While $DK_{pub}$ and $DC$ are considered confidential to hide the device identity from potential attackers, $RK_{pub}$ is considered public. 
The $DC$ and the $DK_{pub}$ are stored on the device and accessible to the worker. 

\subsubsection{Proving}
The proving is executed using the proving key $ZK_{prove}$ in the registration module of the worker that registers the device on the blockchain. 
The proving takes the device certificate $DC$ and the device public key $DK_{pub}$ as private input and the public root key $RK_{pub}$ as public input and returns a non-disclosing device handle: $Prove(ZK_{prove},DK_{pub},RK_{pub}, DC) \rightarrow DH$.
Proving comprises two steps: 
\begin{itemize}
    \item First, the device certificate is verified using the root public key: $VerifyCert(RK_{pub},DC,RK_{pub}$).
    \item Second, a non-disclosing device handle is created as the commitment to the device's public key: $Commit(DK_{pub}) \rightarrow DH$. 
\end{itemize}
On successful execution, the resulting proof $\pi$, the $DH$, and the public input $RK_{pub}$ are wrapped into a blockchain transaction and submitted to the verifier contract. 

\subsubsection{Verification}
The verification is executed by the verifier contract using the verification key $ZK_{verif}$. 
It takes the proof, the device handle, and the public root key as inputs: $Verify(ZK_{verif},\pi, DH, RK_{pub})$.
This assures that the computation has correctly been executed. 
In a second step, the contract asserts that the root key $RK_{pub}$ has been used for proving by comparing it against the root key $RK'_{pub}$ anchored by the task initiator: $Assert(RK_{pub},RK'{pub})$
If successful, the device handle $DH$ is registered on the registration contract and can now be used to verify the data authenticity of the learning workflow.

\begin{figure*}[t]
    \centering
    \includegraphics[width=0.85\textwidth]{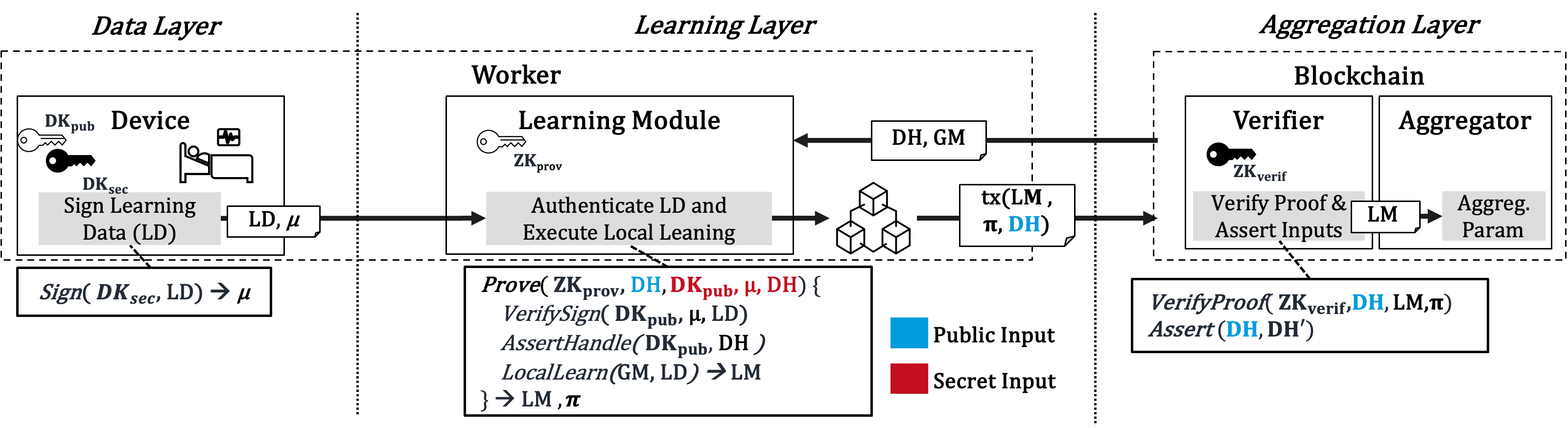}
    \caption{Iterative \textbf{Learning Workflow} for Non-disclosing Verification of Model Integrity and Data Authenticity.}
    \label{fig:learning_workflow}
\end{figure*}

\subsection{Iterative Learning Workflow}
\label{sec:learning_workflow}
The learning workflow is executed in each cycle of the federated learning system by each worker. It reaches from the certified device and to the aggregation contract and is initiated once the global model of the current cycle is available on- chain.

\subsubsection{Attestation}
The data source produces the learning data $LD$ in a predetermined batch size and attests to it by signing the batch using the device secret key: $Sign(DK_{sec},LD) \rightarrow \mu$
As will further be explained in Section~\ref{sec:Implementation}, we assume that signature is created on a commitment to the learning data $comm(LD)$ for efficiency reasons and, as such, $\mu$ additionally contains $comm(LD)$.
The learning data $LD$ and the signature $\mu$ are then provided to the worker's learning module. 

\subsubsection{Proving}
The proving is executed in the learning module using the proving key $ZK_{prov}$ and, in addition to the local parameter update comprises authenticity-related operations. 
The worker obtains the learning data $LD$ and the signature $\mu$ from the device, the current global model $GM$ from the aggregator contract, the device handle $DH$ from the registration contract, and executes the prove on these inputs: $Prove(ZK_{prove},DH,DK_{pub},\mu,DH)$. 
The proving logic consists of three steps and returns the updated parameters as local model $lm$ and the proof $\pi$.
\begin{itemize}
    \item First, the signature $\mu$ over the learning data is verified using the device public key: $VerifySign(DK_{pub}, \mu, LD)$. For that, the commitment of the learning data must be recreated and compared to the one contained in $\mu$.
    \item Second, the mapping of the device handle and the device public key is asserted: $Assert(DK_{pub},DH)$.
    For that, the commitment to the $DK_{pub}$ is created and compared to the $DH$ obtained from the blockchain. 
    \item Third, the local model $LM$ is computed using the learning data and the global parameters as inputs: $LocalLearn(GM,LD) \rightarrow LM$.
\end{itemize}
Upon successful proving, the newly computed local model $LM$, the proof $\pi$, and the public input $DH$ are submitted as a blockchain transaction to the verifier contract. 

\subsubsection{Verification}
The verification is executed through the verifier contract. 
First, the proof is verified using the verification key $ZK_{verif}$ based on arguments contained in the blockchain transaction: $VerifyProof(ZK_{verif},DH,LM,\pi)$.
Then, the contract asserts that a valid device public key has been applied during proving by checking if the device handle is registered at the registration contract: $Assert(DH,DH')$.
If successful, the verified local model is applied for aggregation by the aggregator contract.

\section{Implementation}
\label{sec:Implementation}
\begin{table*}[t]
\tiny
\caption{Average execution time and memory consumption for varying batch sizes \label{tab:exp2}}
\resizebox{\textwidth}{!}{%
\begin{tabular}{l|c|c|c|c|c|c|c|c|c} 
\hline
\multicolumn{1}{c|}{}   & Registration                                                                  & \multicolumn{2}{c|}{Batch size: 10} & \multicolumn{2}{c|}{Batch size: 20} & \multicolumn{2}{c|}{Batch size: 30} & \multicolumn{2}{c}{Batch size: 40}  \\ 
\cline{3-10}
                                       & workflow                                                                      & {[}12]    & Proposed    & {[}12]    & Proposed    & {[}12]    & Proposed    & {[}12]    & Proposed    \\ 
\hline\hline
Compile Time [sec]                     & 5.288                                                                         & 61.816    & 77.099      & 169.564   & 193.236     & 323.723   & 351.811     & 518.96    & 545.239     \\
Constraints [count]                    & 150,696                                                                       & 2,236,596 & 2,398,897   & 4,518,156 & 4,690,549   & 6,799,716 & 6,981,385   & 9,081,276 & 9,273,517   \\
Compile Output Size [GB]               & 0.307                                     & 0.960     & 1.292       & 1.938     & 2.286       & 2.917     & 3.279       & 3.895     & 4.273       \\ 
\hline
Setup Time [sec]                       & 4.135                                                                         & 43.784    & 47.336      & 84.857    & 88.099      & 119.889   & 122.897     & 170.191   & 174.361     \\
Proving Key Size [GB]                   & 0.064                                                                        & 0.981     & 1.032       & 1.976     & 2.031       & 2.703     & 2.761       & 3.967     & 4.028       \\ 
Verification Key Size [Byte]              & 5689                                                                        & 31437     & 35701       & 31437     & 35701       & 31437     & 35701       & 31437     & 35701       \\ 
\hline
Witness Computation Time [sec]         & 1.975                                                                         & 7.888     & 9.991       & 15.973    & 18.161      & 24.289    & 26.328      & 32.802    & 34.565      \\
Computed Witness Size [GB]             & 0.00594                                                                       & 0.089     & 0.095       & 0.179     & 0.186       & 0.270     & 0.278       & 0.361     & 0.369       \\ 
\hline
Proof Generation Time [sec]            & 3.565                                                                         & 28.484    & 31.362      & 56.417    & 59.656      & 73.256    & 76.77       & 113.769   & 117.69      \\
Generated Proof Size [Byte]            & 2699                                                                          & 14317     & 16241       & 14317     & 16241       & 14317     & 16241       & 14317     & 16241       \\ 
\hline
Proving Time [sec]                     & 5.540                                                                         & 36.518    & 42.573      & 72.434    & 79.616      & 97.548    & 106.115     & 146.574   & 155.294     \\ 
\hline
Total Execution Time [sec]             & 11.215                                                                        & 38.780    & 45.032      & 74.759    & 82.511      & 99.923    & 108.77      & 149.029   & 159.01      \\ 
\hline
Verification Cost [GAS]                & 559376                                                                        & 2315527   & 2597842     & 2363812   & 2687877     & 2303386   & 2638482     & 2306273   & 2614425     \\
\hline
\end{tabular}
}
\end{table*}

We implemented the system as a proof-of-concept (PoC)\footnote{https://github.com/eechn/End-to-End-Verifiable-Decentralized-Federated-Learning}.
To establish comparability of the prototype with the state-of-the-art, our implementation builds upon and extends the reference implementation of~\cite{advancing_heiss_2022}. 
The PoC comprises software components for attestation, i.e., CA and devices, proving, i.e., learning and registration module, and smart contracts. 

\textbf{Attestation with CA and Devices:}
We implemented the CA and the device as web servers allowing for decoupled provisioning of data to the proving modules. 
For simulation purposes, we omitted the intermediate certificate storage on the device. 
As a signature algorithm, we decided on the EdDSA using the alt\_bn128 (babyjubjub) curve that allows for verification in the zkSNARKs-based proving environment. 
For efficiency reasons, we decided to create the signature on hashed inputs. 
While not critical for signing small certificate attributes, hashing becomes relevant for signing large batches of learning data. 
As a hash algorithm, we decided on Poseidon~\cite{grassi2021poseidon}, given its high performance for zkSNARKs-based usage. 
To create these signatures, we leveraged the PyCrypto\footnote{https://github.com/Zokrates/pycrypto} library supporting the creation of signatures in a zkSNARK-friendly format. 
Poseidon hashes were pre-computed with ZoKrates.

\textbf{Proving Modules:}
Building upon previous work, we implemented the same simple feedforward neural network as in~\cite{advancing_heiss_2022}. 
The matrix calculation is executed on the hidden layer based on the input arguments, the prediction is determined through an argmax function, and the loss function is implemented as the mean squared error. 
The local model consists of the weights and biases as learning parameters. 
To implement zkSNARKs-based proving in the learning and registration module, we used the ZoKrates toolbox and language (compare Section~\ref{sec:zksnarks}) with Groth16 proving scheme and the alt\_bn28 (babyjubjub) elliptic curve. 
The ZoKrates language is compiled into the $ecs$ that is used for the one-time setup and the iterative witness computation and proof generation. 

\textbf{Smart Contracts:}
The verification contract for each zkSNARK program was generated using the ZoKrates toolbox which directly integrates the verification key into the verification logic of a Solidity contract. 
The aggregation contract was adopted from previous work given the same FL logic. 
It maintains two versions of the weight vector and the bias matrix, one used to implement incoming updates and one representing the latest learning state. 
After a learning cycle terminates (based on block time), the latest learning state changes to the updated one.
The registration contract contains a key-value registry of identifiers and device handles that is filled after successful certificate verification and requested for each learning update. 
The contracts are deployed on a virtual Ethereum blockchain using Truffle and Ganache. 

\textbf{Challenges:}
By implementing the zkSNARKs-based logic we faced some challenges:
Firstly, the ZoKrates language only supports arithmetic operations on positive integer values, however, learning data often contains negative and floating point numbers. Data can be translated into positive integers by upscaling the input data which, however, can result in integer overflows. Taking this account, we adopted the strategy of previous work ~\cite{advancing_heiss_2022} and reimplemented arithmetic operations in the ZoKrates program. Next, we were to decide if the authenticity proofs were heavy on the signature algorithm, i.e., signing large plain learning data, or on the hash function, i.e., hashing learning data prior to signing. We observed the best performance on large batch data in a hash-heavy approach using the Poseidon hash function in Merkle trees. A root hash was constructed from the batch data and signed during attestation. The same root hash was then recreated inside ZoKrates to verify the signature.

\section{Evaluation}
\label{sec:Evaluation}
We conducted experiments on our implementation for both workflows. 
While the registration workflow is evaluated on its own, the learning workflow is evaluated in comparison with experiments of~\cite{advancing_heiss_2022}. 

\subsection{Experimental Data and Setup}
\textit{Datasets} for registration and learning were selected in accordance with the system model. As inputs for the registration workflow, we used an EdDSA public key as a certificate attribute and input to the registration workflow. For learning workflow, we used the data, obtained from the UC Irvine Machine Learning Repository\footnote{https://archive.ics.uci.edu/dataset/256/daily+and+sports+activities} that consists of sensor data generated by wearables. It was collected at 5-minute intervals from 8 subjects performing 19 pre-defined activities. As in~\cite{advancing_heiss_2022}, we condensed the original dataset containing 45 features and 19 prediction classes to a dataset of 9 features and 6 classes to match the input vector of the learning model and the size of the hidden layer's weights and biases. 

\textit{Iterations} of the experiments depend on the workflow type.
Given the one-time registration workflow, there was no need for iterative executions. Instead, the learning was repeated 300 times for each batch size using different batch sizes of 10, 20, 30, and 40 elements. We measured the system performance with a single worker on a single machine.

\subsection{Results}
\textit{Constraint numbers} of the $ecs$ used in the proving module demonstrate the complexity of the proving logic. 
As depicted in Table~\ref{tab:exp2}, the proving registration is considerably cheaper than the proving the parameter update. 
However, our solution added only a relatively small number of constraints compared to the reference implementation without authenticity proofs. 

\textit{Execution Times} of the proving may impact the duration of the learning cycles if it takes too long. 
For the registration workflow, the total proving took 5.5403 seconds. 
Added to a negligible certificate creation of 0.0816 seconds and a verification time of 2.822 seconds, the overall execution time for the device registration process is around 10 seconds which we deem acceptable for a one-time execution. Compilation and setup took 5.288 sec and 4.135 sec respectively.
For the learning workflow, there was no significant difference in the proving.
The witness and proof creation time took up to 3.921 (Batch size: 40) longer. 
Similar proportional differences can be observed for the compilation and setup which, however, is non-critical given that they are only executed for system initialization. 
Overall, the operational learning workflow takes about 10 seconds longer compared to the reference.

\textit{Transaction Costs} are critical given redundant blockchain transaction processing and expensive consensus protocols. 
Given a constant proof size of zkSNARKs, the verification costs vary depending on the size of the public inputs and the computational results. 
The costs for verifying the registration proof account to 559376 Gas. 
The transaction costs for learning workflow are considerably higher due to the model aggregation. 
However, our implementation only adds a small cost overhead compared to the ones of the reference which may be caused by the additional device handle operation.

\begin{figure}[h]
    \centering
    \includegraphics[width=0.9\columnwidth]{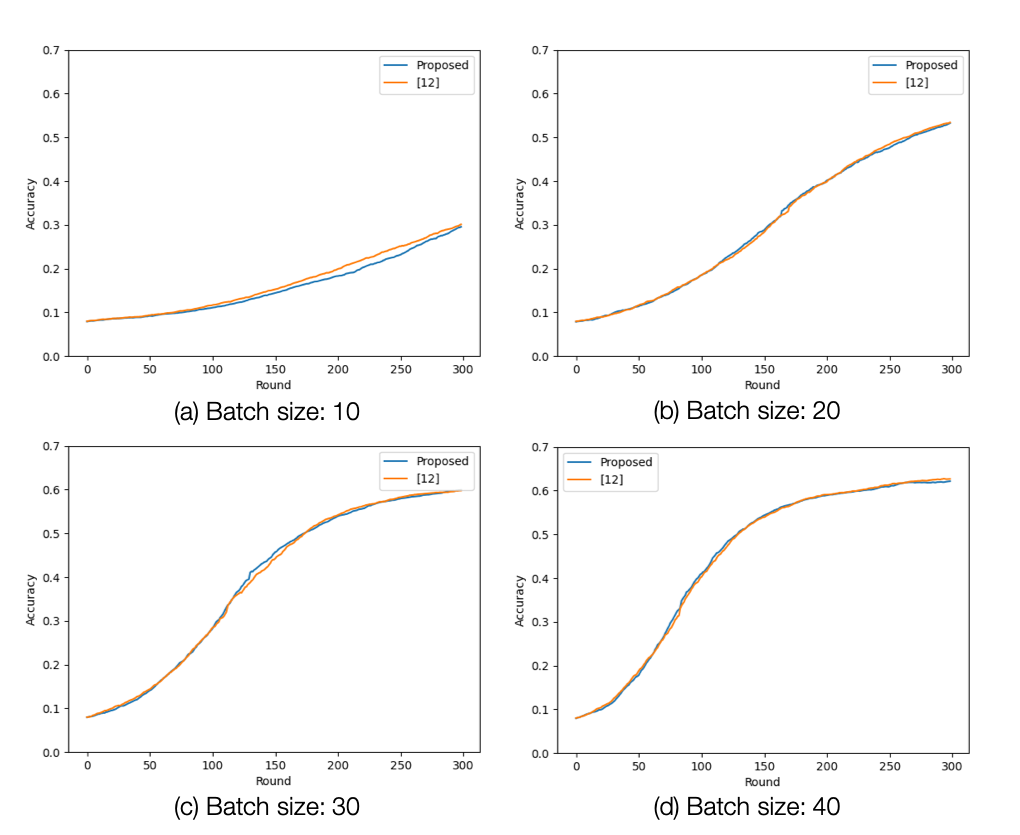}
    \caption{Accuracy for various batch sizes}
    \label{fig:score}
\end{figure}

\textit{Accuracy} of the prediction as the learning outcome is specific to the learning experiments. 
The trend of the scores is shown in Figure ~\ref{fig:score}, across 300 epochs with varying batch sizes.
As the batch size increases, the model achieves higher scores more rapidly and with greater accuracy.
Since both systems utilize the same training algorithm, the difference in learning performance is not prominent. 
A relatively low accuracy score (a maximum accuracy of 0.62 when the batch size was set to 40) is derived since there is only one node participating in federated learning to update the global model. 
However, as the number of workers increases, better scores can be obtained.

\vspace{-0.2cm}
\section{Discussion}
\label{sec:Discussion}
The evaluation demonstrated the practicality of the proposed system workflows. 
In this section, we revisit aspects of attack resistance and discuss opportunities for managing computational overheads and system artifacts. 

\subsection{Security and Trust}

\textit{Workers} are verifiers and potential attackers at the same time. Recalling the verifiability objectives, our system enables workers to verify and protect against attacks on global tasks through blockchains, local tasks through zkSNARKs, and device certificates and learning data through integrated non-disclosing authenticity proofs. Beyond that, lazy workers may resubmit already accepted model updates that would pass the 2PV procedure again. Such replay attacks can be prevented also from external attackers through batch counters that are signed by the devices together with the learning data and checked during proving and verification on-chain. 

\textit{Task Initiator}: 
By executing the setup, a malicious task initiator may corrupt the zkSNARKs setup by keeping the structured reference string (SRS) to create fake proofs. To remove trust assumptions from the setup procedure, secure multiparty computation (sMPC) has been established to make the setup verifiable across involved parties~\cite{ben2015secure}. By applying sMPC to our system, workers can verify the setup procedure and trust assumptions can be removed from the task initiator. 

\textit{Certificate Authority}: 
With the responsibility of issuing device certificates that confirm the well-functioning of the devices, the CA bears trust assumptions. For one, the CA could execute a Sybil attack by issuing fake certificates to faulty or non-existent devices. Furthermore, the CA could hinder well-functioning devices from participating by refusing certificate issuance. Similar effects could be achieved through man-in-the-middle attacks where the communication channel between CA and device is intermitted through an attacker faking messages in both directions. Addressing such security risks research has been conducted on Decentralized Public Key Infrastructure~\cite{9119673, 10.1007/978-3-030-02641-7_14}. We consider a system extension towards DPKI as a possible future work.

\subsection{Managing Computational Overheads}
ZkSNARKs-based proving and blockchain-based aggregation introduce verifiability but also severe computational overheads that can represent obstacles for practical deployments. 

To reduce transaction costs of the blockchain, further global tasks, like the aggregation, could be outsourced to an off-chain node as a verifiable off-chain computation, similar to~\cite{zkfl_knottenbelt_2023}.
While preserving verifiability, such an extension, however, would introduce novel challenges, e.g., regarding the system’s liveness.  
Furthermore, the choice of the blockchain consensus protocol has a significant impact on the transaction costs but strongly depends on the requirements of the application scenario, e.g., permissioned versus permissionless settings. 

A promising approach for optimizing the proving is the usage of a recursive proof system like Nova~\cite{10.1007/978-3-031-15985-5_13}. 
Such systems help to execute larger computational tasks on machines with constrained memory. 
Instead of executing a large batch at once, the proving is executed recursively on smaller chunks resulting in a single proof that can be verified on-chain. 
Furthermore, proving costs can be reduced through circuit optimization resulting in smaller $ecs$ or by leveraging specialized hardware that implements the circuits closer to the processor.

\subsection{Distributing Proving Artifacts}
The proving key $ZK_{prove}$ and the $ecs$ are required by the workers to execute the proving. However, as can be seen in Table~\ref{tab:exp2}, they are large files that can and should not be stored on a storage-constrained blockchain. To make them accessible to the workers, we propose that the $ZK_{prove}$ and $ecs$ of the registration and learning modules are provided through IPFS to preserve integrity, thereby unburdening the blockchain storage while preserving integrity. To further guarantee availability, IPFS can be extended through a protocol like Filecoin that provides storage incentives to decentralized nodes. Using the content addressable storage pattern~\cite{10.1007/978-3-319-67262-5_1}, the artifacts can be linked from the verification contract while keeping them off-chain and can be retrieved faster. Additionally, we propose to provide the module applications in a human-readable format, e.g., the ZoKrates high-level language, to enable workers to understand the proving logic and to reproduce the $ecs$.

\section{Conclusion}
\label{sec:Conclusion}
We proposed a first system for end-to-end verifiable decentralized FL that extends state-of-the-art systems through non-disclosing authenticity proofs, thereby adding verifiability to the data source. 
As the core of the system, we introduced a two-step proving and verification method that, first, verifies authenticity proofs from the data layer on the learning layer, and second, verifies zkSNARKs from the learning layer on the blockchain-based aggregation layer. 
We applied this procedure to a registration and learning workflow enabling end-to-end verifiability between trusted data sources and the blockchain without revealing confidential information. 
Experimental results demonstrate the feasibility of the proposed system with only marginal overhead compared to the reference implementation. 

In future work, we aim to incorporate aspects discussed in the ~\ref{sec:Discussion}, especially enhancements of the system performance through recursive proof systems and further outsourcing of global FL tasks.



\bibliographystyle{IEEEtran}
\bibliography{references}

\end{document}